\pdfoutput=1

\documentclass[11pt]{article}

\usepackage[]{naacl2021}

\usepackage{times}
\usepackage{latexsym}

\usepackage[T1]{fontenc}

\usepackage[utf8]{inputenc}

\usepackage{microtype}
\usepackage{graphicx}
\usepackage{multirow}

\title{OCID-Ref: A 3D Robotic Dataset with Embodied Language for Clutter Scene Grounding}

\author{
Ke-Jyun Wang\thanks{$^*$ Equal contribution.} \qquad Yun-Hsuan Liu\footnotemark[1] \qquad Hung-Ting Su \qquad Jen-Wei Wang \\  {\bf Yu-Siang Wang \qquad Winston H. Hsu \qquad Wen-Chin Chen}
\\National Taiwan University
}

\begin{document}
\maketitle


\begin{abstract}
To effectively apply robots in working environments and assist humans, it is essential to develop and evaluate how visual grounding (VG) can affect machine performance on occluded objects. However, current VG works are limited in working environments, such as offices and warehouses, where objects are usually occluded due to space utilization issues. In our work, we propose a novel OCID-Ref dataset featuring a referring expression segmentation task with referring expressions of occluded objects. OCID-Ref consists of 305,694 referring expressions from 2,300 scenes with providing RGB image and point cloud inputs. To resolve challenging occlusion issues, we argue that it's crucial to take advantage of both 2D and 3D signals to resolve challenging occlusion issues. Our experimental results demonstrate the effectiveness of aggregating 2D and 3D signals but referring to occluded objects still remains challenging for the modern visual grounding systems. OCID-Ref is publicly available at \url{https://github.com/lluma/OCID-Ref}

\end{abstract}

\begin{figure}[t]
\begin{center}
\includegraphics[width=1.0\linewidth]{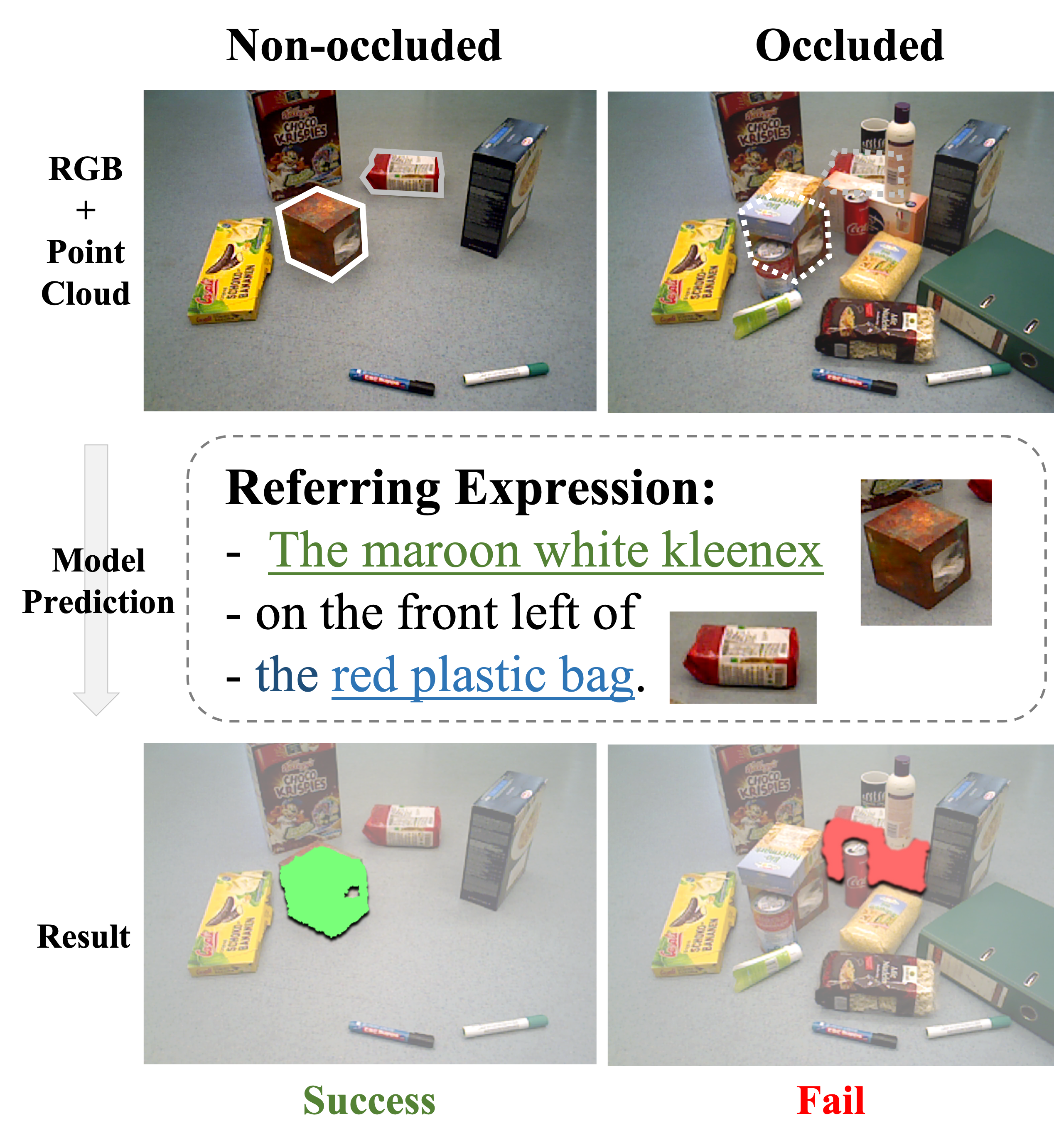}
\end{center}
\caption{\label{fig:dataset}
A hard case where visual grounding (VG) network fails to predict the occluded object in clutter scene. Our dataset provides more such cases than other datasets, which are commonly seen in the working spaces, like offices and warehouses.
}
\end{figure}


\section{Introduction}
Visual grounding (VG), which aims to locate the object according to a structured language query, is a crucial task in natural language processing (NLP), computer vision (CV), and robotics. Recent VG studies most focus on \textit{web-crawled images} such as  \cite{KazemzadehOrdonezMattenBergEMNLP14,krishnavisualgenome,mao2016generation,referitgameeccv2016}. However, VG for human-robot interaction (HRI) is less explored. Most of the images in existing VG datasets are people and daily necessities, e.g., RefCOCO contains mainly persons, cars, and cats, which are separated and therefore easier to detect. Nevertheless, working spaces such as offices or warehouses, where robots are usually applied to assist works, are usually crowded, and objects are overlapped with each other to utilize space better. Therefore, objects in working environments are often occluded and hard to detect.


Previous work \cite{ralph2005human} suggested that a system that uses language for human-computer interaction can help non-professionals instruct robots to complete technical work and collaborate. Recent research pointed out that VG plays an important role in HRI. \cite{shridhar2018interactive} utilized VG to resolve ambiguity in grasping tasks. \cite{matuszek2018grounded} studied how the robot learns about objects and tasks in an environment via nature language queries. Therefore, explicit language instructions and good referring (grounding) expressions are pivotal in human-robot interaction and improve communication between non-expert humans and robots.

Some efforts have been made to collect VG datasets. RefCOCO \cite{referitgameeccv2016} and Cops-Ref \cite{chen2020copsref} utilize web-crawled images and manually label language expressions. A limitation is that images alone do not provide precise position cues, which are essential for various downstream robotic tasks such as grasping. A recent work, Sun-Spot \cite{mauceri2019sun}, utilizes a depth channel for object detection and referring expression segmentation tasks. Another existing dataset, ScanRefer \cite{chen2019scanrefer}, uses more accurate multi-view point clouds for 3D signals. However, both Sun-Spot and ScanRefer do not address \textit{occlusion} issues, which is ordinary in working spaces and more challenging due to more compositions of shapes of each object. As shown in figure \ref{fig:dataset}, when an object \textit{(the red plastic bag)} is blocked in an occluded environment, the shape of the object could be deformed and increase VG difficulty.

\begin{figure}[t]
\begin{center}
\includegraphics[width=1.0\linewidth]{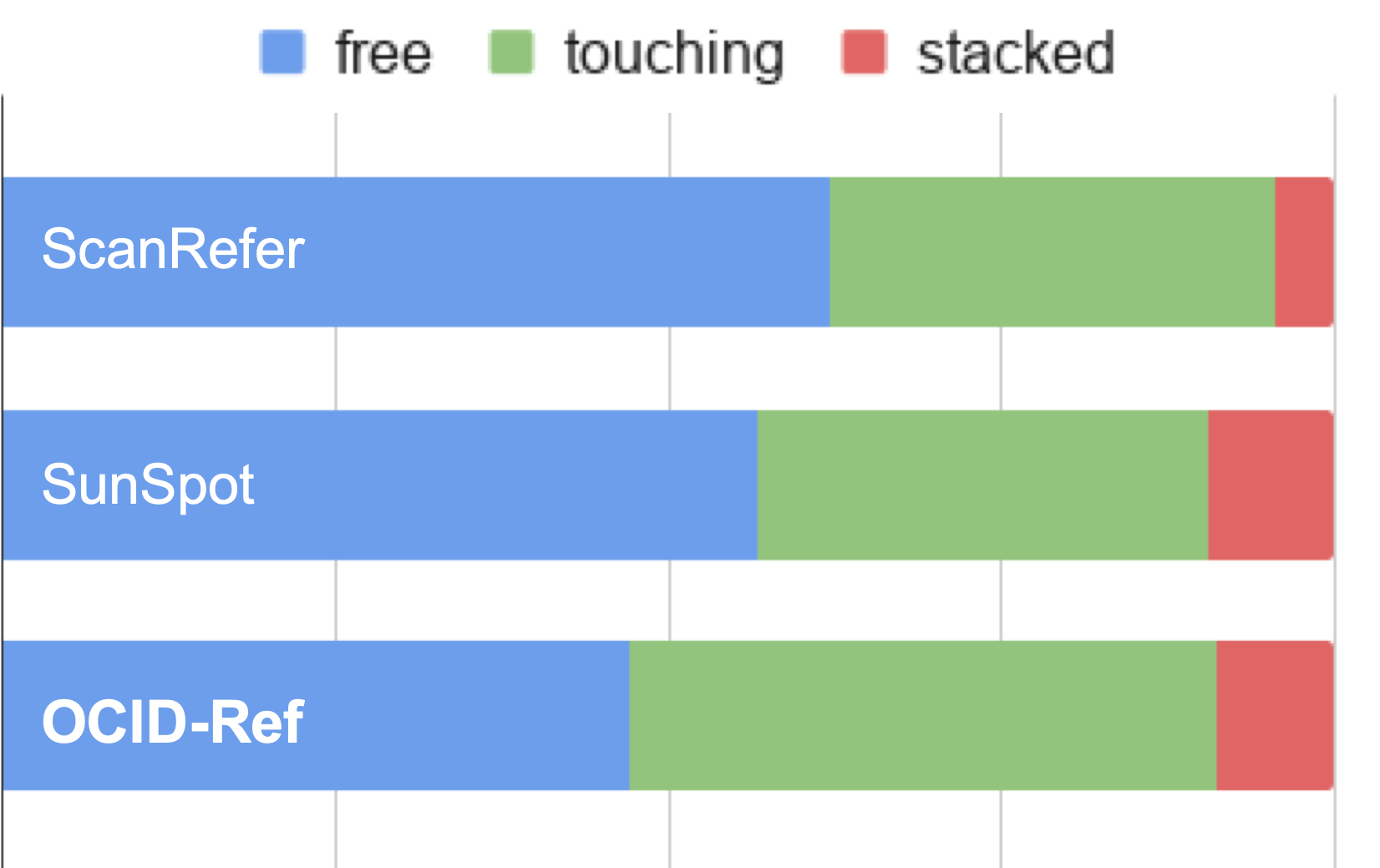}
\end{center}
\caption{\label{fig:clutter-lev}
\textbf{Clutter level ratio.} Our proposed dataset, OCID-Ref, has more challenging and practical examples (touching and stacked) than ScanRefer~\cite{chen2019scanrefer} and SunSpot~\cite{mauceri2019sun} for visual grounding task.
}
\end{figure}

Observing this, we propose a novel OCID-Ref dataset with two key features: (1) For each scene, we utilize both RGB image and point cloud to provide multi-modal signals for learning system development. (2) OCID-Ref scenes have higher clutter level compared to existing datasets, as shown in figure \ref{fig:clutter-lev}. Hence, the model capability for resolving challenging occlusion issues could be evaluated. To the best of our knowledge, OCID-Ref is the only existing dataset supporting the above features, and therefore allows VG task in grasping scenario. 

Experimental results demonstrate that occluded scenes are more challenging to modern VG baselines. We observe 27\% to 34\% performance drops on referring expression segmentation tasks. Also, utilizing 3D information continually improves performance across all clutter levels. Furthermore, fusing 2D and 3D features reach the best performance on all clutter levels. We suggest that OCID-Ref dataset could pave a new path for VG research in HRI and benefit the research community and application developments. 

\begin{table*}
\begin{center}
\begin{tabular}{ccccccc}
\hline
 & \shortstack{\# Scenes/\# Images} & \shortstack{\# Obj. Cat.} & \shortstack{Dis. Score} & \shortstack{Data Format} &
 \shortstack{\# Expressions} & \shortstack{AvgLen}
 \\
\hline
RefCOCO   & 19,994        & 80        & 4.9        & RGB        & 142,210        & 3.5 
\\
Cops-Ref  & 75,299        & 508       & 20.3       & RGB        & 148,712        & 14.4                                                       
\\
Sun-Spot  & 1,948         & 38        & 2.46       & RGB-D      & 7,990          & 14.04                                                        
\\
ScanRefer & 703           & 18        & 4.64       & 3D         & 46,173         & 17.91 
\\
OCID-Ref  & 2,300         & 58        & 3.36       & 3D         &  267,339       & 8.56
\\ 
 \hline
\end{tabular}
\end{center}
\caption{\label{tab:dataset-stat}
Statistic comparison of previous 2D, RGB-D, 3D referring datasets and the OCID-Ref  in terms of the number of scenes or images (\#Scenes/\#Images), number of object categories(\#Obj. Cat.), Distractor score(Dis. Score), Data format, number of expressions (\#Expressions), and average lengths of the expressions(AvgLen). Our OCID-Ref is the first dataset featuring both 3D signals and object occlusion, which are both crucial for visual grounding for HRI.
}
\end{table*}

\section{Dataset and Task}
To open up a new way for VG research in HRI, we collect a novel OCID-Ref dataset by the following steps: (1) We leverage a robotic object cluttered indoor dataset, OCID  \cite{DBLP:conf/icra/SuchiPFV19}, which consists of complex clutter-level scenes with rich 3D point cloud data and the point-wise instance labels for each occluded objects. (2) We manually annotate fine-grained attributes and relations such as color, shape, size relation or spatial relation. (3) We generate referring expressions based on annotated attributes and relations with a similar scene-graph generation system from \cite{yang2020graph-structured} and \cite{Chen_2020_CVPR}. In this section, we will describe more details on our data collection and the scene-graph generation method we adopt to generate the referring expressions.

\subsection{Data Collection}
A proper dataset to evaluate and develop VG models in a working environment requires two properties: (1) cluttered scenes and (2) 3D signals. To point out the important of these two properties, we conduct a pilot experiment of grasp detection\footnote{See Appendix A for details}.  We observe that using 3D cues significantly boosts performance,  the geometric features extracted from point cloud data benefit the robots on visual perception (e.g., object grasping or object tracking). Also, we see a severe performance drop in occluded scenes.

Therefore, to provide scenes with occluded objects to develop and evaluate learning systems, we leverage an existing robotic 3D dataset, OCID \cite{DBLP:conf/icra/SuchiPFV19}, which has higher clutter level scenes and sequential object-level scenes that help robots better understand the instance difference between two subsequent scenes.

Hence, we choose OCID as our original dataset, and extend it with extra semantic annotations such as attributes (e.g., color, texture, shape) and relations (e.g., color relation, spatial relation, etc.) for all the objects in dataset. We design an online web-based annotation tool to collect these extra labels, and dispatch the labeling tasks over the annotation specialists from a professional data service company. Additionally, we ensure each task is randomly assigned to three trained workers and verified by one checker. The overall tasks take around two months to finish. 

\subsection{Referring Expression Generation} 
Gathering the labels we annotated and following the method from the scene-graph based referring expression generation system. In detail, first, we build up the scene graph for each scene in OCID-Ref, and the nodes and edges in the graph represent the attributes and relations, respectively. Second, we design several textual templates (\autoref{tab:dataset-templates}) to have various sentence structures. Third, we leverage the conventional incremental algorithm\cite{DALE1995233} and functional programs to generate reasonable REs. That is, we add attributes and relations into our conditional set until it conforms with the specific unambiguous condition. Finally, we generate the total of 305,694 referring expressions with an average length of 8.56, and for details, there are an average of 14.71 expressions per object instance and 113.07 expressions per scene.

\begin{table}
\begin{tabular}{cc}
\hline
 & \shortstack{Exemplar Templates} 
 \\
\hline
\multirow{2}{4em}{Common Sentence} & The <Attr> <Obj>. \\ 
& The <Attr> object / item. \\ 
\hline
\multirow{3}{4em}{Relational Sentence} & The <Obj> <Rel>. \\ 
& The <Attr> <Obj> <Rel>. \\ 
& The <Attr> <Obj1> <Rel> <Obj2>. \\ 
\hline
\end{tabular}
\caption{\label{tab:dataset-templates} 
Show the exemplar templates we use to generate the free-form referring expressions.
}
\end{table}

\subsection{Dataset Statistics}
OCID-Ref uses the same scenes as OCID, containing 2D object segmentation and 3D object bounding boxes for 2300 fully built-up indoor cluttered scenes. Each object is associated with more than 20 relationships with other objects in the same scene, including 3D spatial relations, 2D spatial relations, comparative color relations, and comparative size relations. 
Table \ref{tab:dataset-stat} shows the basic statistic comparison of the previous 2D, RGB-D,3D referring datasets and the OCID-Ref. 
To evaluate the difficulty of REC, we follow Cops-Ref to calculate the number of candidate objects of the same categories as the target object(Distractor score) for all scenes. Though there are only 3.36 same candidates in an average of OCID-Ref, lower than 4.64 of ScanRefer, we attribute this difference to the dataset characteristic that our scenes are components of one by one sequence with few objects in the first few scenes. To evaluate the referring performance from no clutter to dense clutter scenes, we follow OCID to separate the scenes into three cluttered levels, free, touching, and stacked, from clearly separated to physically touching to being on top of each other. We also split the val split of ScanRefer into three clutter level.



\begin{table*}
\begin{center}
\begin{tabular}{cccccc}
\hline 
 & All & Free & Touching & Stacked & Decrease rate(free->stacked) $\downarrow$ \\
\hline
2D         & 0.512 / 0.501 & 0.673 / 0.660 & 0.450 / 0.496 & 0.450 / 0.433 & 33.23\% / 34.29\% \\
3D         & 0.588 / 0.580 & 0.745 / 0.751 & 0.589 / 0.583 & 0.507 / 0.489 & 31.97\% / 34.92\% \\
Fusion      & \textbf{0.634 / 0.637} &  \textbf{0.763 / 0.769} &  \textbf{0.64 / 0.651} &  \textbf{0.551 / 0.540} & \textbf{27.71\% / 29.75\%} \\
\hline
\end{tabular}
\end{center}
\caption{\label{tab:performance}
Referring expression segmentation performance (pr@0.25) on OCID-Ref. The performance is negatively associated with the clutter level, indicating that the occlusion could make the VG task more challenging. Also, leveraging 3D signals enhances the overall performance on all clutter levels.
}
\end{table*}

\section{Experiments}
We conduct referring expression segmentation experiments on our collected OCID-Ref dataset and ScanRefer \cite{chen2019scanrefer} dataset. We compare different modalities, clutter levels, and regular expression lengths and provide a comprehensive analysis to pave a new path for future research. We also conduct the grasp experiment using different modality data as input, and the details are described in Appendix A.

\begin{figure}[t]
\begin{center}
    \includegraphics[width=0.8\linewidth]{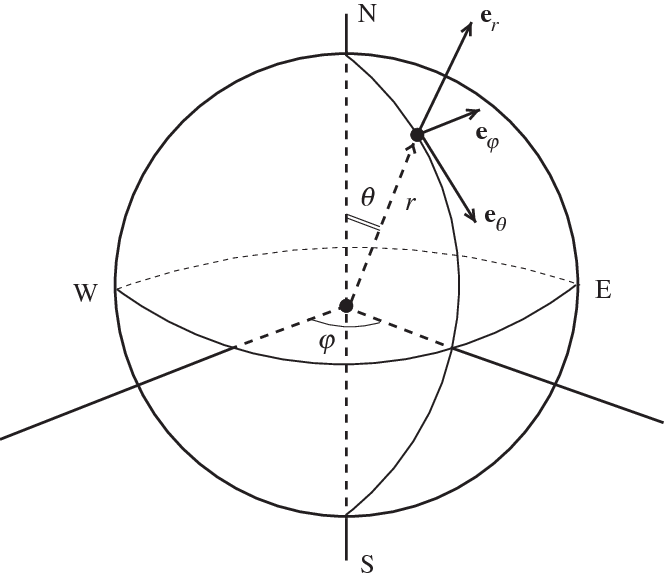}
\end{center}
\caption{\label{fig:rel-def}
Compute the angles related to 3D relation on spherical coordinates.
}
\end{figure}

\begin{equation} 
    \label{eq:1}
    R(\theta,\phi) 
    = \left\{ 
        \begin{array}{l} 
            4\;\;\; if\; \theta < 15\\
            5\;\;\; if\; \theta > $-15$ \\
            16 + r(\phi)\;\;\; if\; $-65$ < \theta < $-15$ \\
            8 + r(\phi)\;\;\; if\; $15$ < \theta < $65$ \\
            r(\phi)\;\;\; otherwise \\
        \end{array}
    \right. 
\end{equation}

\begin{equation}
    \label{eq:2}
    r(\phi) = 6 + \left(\frac{\phi + 22.5}{45}\right) 
\end{equation}

\begin{figure*}[t]
\begin{center}
\includegraphics[width=1.0\linewidth]{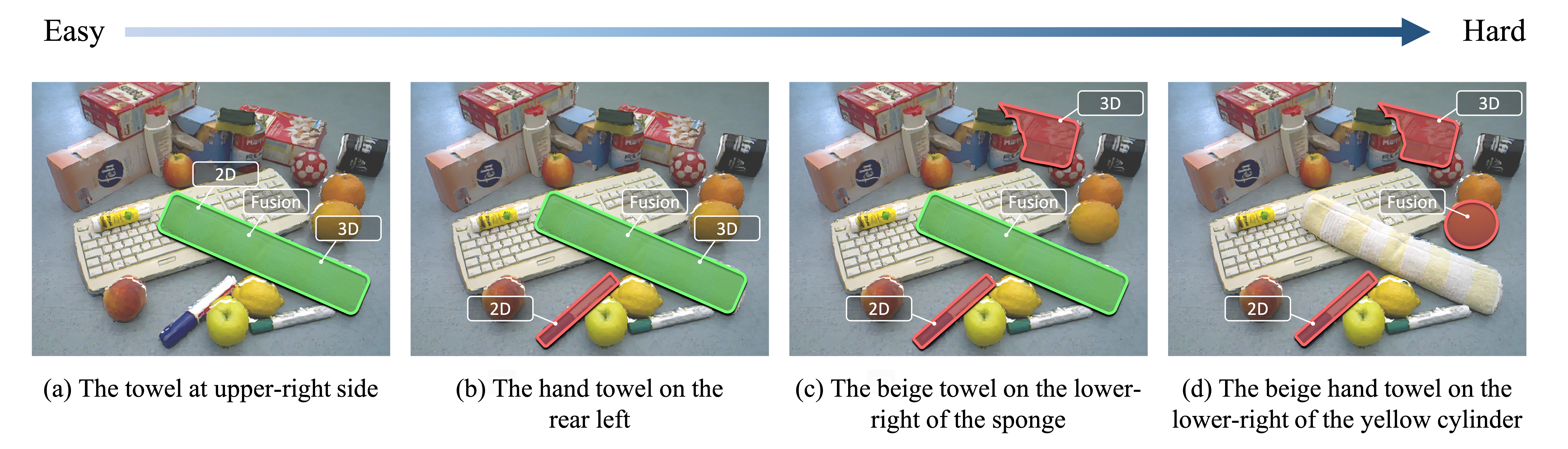}
\end{center}
\caption{\label{fig:qal}
Qualitative results from 2D, 3D, and the fusion methods. Predicted masks with an IOU score higher than 0.25 are marked in green, otherwise in red. Examples are tested in the same cluttered scene with referring expressions in different difficulty levels. Fusion method produces better results than 2D and 3D method.
}
\end{figure*}

\subsection{Setup}
\paragraph{Baseline} We run our experiments with a modern graph-based DGA \cite{yang2019dynamic} model. We compare 2D (RGB), 3D (point cloud) and 2D+3D input signals. 
\paragraph{Feature Extraction} For 2D inputs, we use ResNet-101 based Faster-RCNN as our 2D feature extractor and pre-train the extractor on OCID to extract the ROI features from the pool5 layer as the 2D visual features, and use the original DGA's settings for node feature and edge feature on the graph. For 3D inputs, we utilize point-wise features extracted from PointNet \cite{qi2016pointnet} as the 3D version of the visual feature for each node in the graph. Also, we change the box information from 2D to 3D with box center, box bounds, and box volume. The relations for the edges are modified with 3D relationships between objects instead of 2D relationships. Figure \ref{fig:rel-def} and equation \ref{eq:1}, \ref{eq:2} shows how we compute the angles related to 3D relation on spherical coordinates. 




\paragraph{2D and 3D Fusion} To utilize advantages from both 2D and 3D signals, we implement a handy fusion module. We take max-pooling on the point features to aggregate them into a global scene feature and concatenate it to the 2D visual feature as a new visual feature for each object instance. Afterward, we fuse the box information into (2D box center, 2D box bounds, 2D box area, 3D box center, 3D box bounds, 3D box volume) to preserve the location information from two distinct coordinates. The edge representation is defined as the same as the 3D version.

\paragraph{Evaluation Metric} We use Acc@0.25IoU as our metric to measure the thresholded accuracy where the positive predictions have a higher intersection over union (IoU) with the ground truths than the thresholds.

\subsection{Quantitative Analysis}
\paragraph{Clutter Levels} Table \ref{tab:performance} compares 2D (RGB), 3D (point cloud) models and Fusion model performance on OCID-Ref dataset. Obviously, all models struggle against the highly occluded \textit{stacked} subset (Fourth column). The 27 to 34 \% of performances drop from free to stacked subset indicates that occlusion, which occurs in working environments, is a challenge for modern VG models. Table \ref{tab:clutter-level-scanrefer} shows model performance on ScanRefer dataset, and the result is consistent with OCID-Ref dataset, where stacked performance is dropped from 0.465 to 0.320 for the unique scenario and from 0.198 to 0.131 for the multiple scenario. The results suggest that tackling occlusion is crucial for future research and applications in working environments.

\paragraph{Input Modality} As shown in table \ref{tab:performance}, for single modality models, the 3D model (Second row) constantly outperforms the 2D model (First row) in all clutter levels and indicates that accurate spatial information is crucial. Furthermore, aggregating 2D and 3D signals (Third row) reaches the best performance and suffers less performance drop from free to stacked. Therefore, we suggest future work to explore an effective way to utilize and fuse 2D and 3D signals to tackle our challenging dataset.

\paragraph{Referring Expression Length}
 Table \ref{tab:ablation-length} compares the performance of short (not more than 12 wordpieces) and long (equal or more than 12 wordpieces). We observe that all models perform worse when the expressions are long.

\subsection{Qualitative Analysis}
Figure \ref{fig:qal} shows results produced by 2D, 3D baseline, and the fusion model. First, in figure \ref{fig:qal}-d we discover that all three methods fail when the RE is long and complicated. The fusion method successfully localizes the towel in the scene with 2D and 3D spatial descriptions(refer to figure \ref{fig:qal}-c), while the 3D method has difficulty identifying what is "lower-right." Unsurprisingly, we observe that the 2D method fails on the query with the 3D relation "rear"(refer to figure \ref{fig:qal}-b). 
Figure \ref{fig:qal}-d  also shows the failure cases of the fusion method, indicating that our model cannot handle all spatial relations to distinguish between ambiguous objects. 2D and 3D get better performance when the query RE consisted mainly of the common sentences and relationships regarding the whole scene. The failure case suggests that our fusion and localization module can still be improved to utilize the 2D information better and decrease the 3D features' misuse.

\begin{table}
\begin{center}
\begin{tabular}{cccc}
\hline
& Free  & Touching & Stacked \\
\hline
unique   & 0.465 & 0.407    & 0.32   \\
multiple & 0.198 & 0.179    & 0.131  \\
\hline
\end{tabular}
\end{center}
\caption{\label{tab:clutter-level-scanrefer}
The performance on ScanRefer in differnet clutter level.
}
\end{table}

\begin{table}
\begin{center}
\begin{tabular}{cccc}
\hline
 & 2D & 3D & Fusion \\
\hline
short                         & 0.508                           & 0.592                           & 0.645                               \\
long                          & 0.484                           & 0.562                           & 0.580     
\\
\hline
\end{tabular}
\end{center}
\caption{\label{tab:ablation-length}
Referring expression segmentation performance on different length of the referring expression.
}
\end{table}

\section{Conclusion}
In this work, we propose a novel OCID-Ref dataset for VG with both 2D (RGB) and 3D (point cloud) and occluded objects. OCID-Ref consists of 305,694 referring expressions from 2,300 scenes with providing RGB image and point cloud inputs. Experimental results demonstrate the difficulty of occlusion and suggest the advantages of leveraging both 2D and 3D signals. We are excited to pave a new path for VG researches and applications.

\section{Acknowledgement}
This work was supported in part by the Ministry of Science and Technology, Taiwan, under Grant MOST 110-2634-F-002-026 and Qualcomm Technologies, Inc. We benefit from NVIDIA DGX-1 AI Supercomputer and are grateful to the National Center for High-performance Computing. We also thank Yu-Kai Huang for his insightful suggestion on the figures.

\bibliography{anthology}

\begin{thebibliography}{19}
\expandafter\ifx\csname natexlab\endcsname\relax\def\natexlab#1{#1}\fi

\bibitem[{Calli et~al.(2015)Calli, Singh, Walsman, Srinivasa, Abbeel, and
  Dollar}]{calli2015ycb}
Berk Calli, Arjun Singh, Aaron Walsman, Siddhartha Srinivasa, Pieter Abbeel,
  and Aaron~M Dollar. 2015.
\newblock The ycb object and model set: Towards common benchmarks for
  manipulation research.
\newblock In \emph{2015 international conference on advanced robotics (ICAR)},
  pages 510--517. IEEE.

\bibitem[{Charles et~al.(2017)Charles, Su, Mo, and Guibas}]{qi2016pointnet}
R.~Charles, Hao Su, Kaichun Mo, and Leonidas Guibas. 2017.
\newblock \href {https://doi.org/10.1109/CVPR.2017.16} {Pointnet: Deep learning
  on point sets for 3d classification and segmentation}.
\newblock pages 77--85.

\bibitem[{Chen et~al.(2019)Chen, Chang, and Nie{\ss}ner}]{chen2019scanrefer}
Dave~Zhenyu Chen, Angel~X Chang, and Matthias Nie{\ss}ner. 2019.
\newblock Scanrefer: 3d object localization in rgb-d scans using natural
  language.
\newblock \emph{arXiv preprint arXiv:1912.08830}.

\bibitem[{Chen et~al.(2020{\natexlab{a}})Chen, Wang, Ma, Wong, and
  Wu}]{chen2020copsref}
Zhenfang Chen, Peng Wang, Lin Ma, Kwan-Yee~K. Wong, and Qi~Wu.
  2020{\natexlab{a}}.
\newblock \href {http://arxiv.org/abs/2003.00403} {Cops-ref: A new dataset and
  task on compositional referring expression comprehension}.

\bibitem[{Chen et~al.(2020{\natexlab{b}})Chen, Wang, Ma, Wong, and
  Wu}]{Chen_2020_CVPR}
Zhenfang Chen, Peng Wang, Lin Ma, Kwan-Yee~K. Wong, and Qi~Wu.
  2020{\natexlab{b}}.
\newblock Cops-ref: A new dataset and task on compositional referring
  expression comprehension.
\newblock In \emph{Proceedings of the IEEE/CVF Conference on Computer Vision
  and Pattern Recognition (CVPR)}.

\bibitem[{Dale and Reiter(1995)}]{DALE1995233}
Robert Dale and Ehud Reiter. 1995.
\newblock \href {https://doi.org/https://doi.org/10.1016/0364-0213(95)90018-7}
  {Computational interpretations of the gricean maxims in the generation of
  referring expressions}.
\newblock \emph{Cognitive Science}, 19(2):233--263.

\bibitem[{Jeng et~al.(2020)Jeng, Liu, Liu, Wang, Chang, Su, and
  Hsu}]{jeng2020gdn}
Kuang-Yu Jeng, Yueh-Cheng Liu, Zhe~Yu Liu, Jen-Wei Wang, Ya-Liang Chang,
  Hung-Ting Su, and Winston~H. Hsu. 2020.
\newblock \href {http://arxiv.org/abs/2010.10695} {Gdn: A coarse-to-fine (c2f)
  representation for end-to-end 6-dof grasp detection}.

\bibitem[{Kazemzadeh et~al.(2014)Kazemzadeh, Ordonez, Matten, and
  Berg}]{KazemzadehOrdonezMattenBergEMNLP14}
Sahar Kazemzadeh, Vicente Ordonez, Mark Matten, and Tamara~L. Berg. 2014.
\newblock Referit game: Referring to objects in photographs of natural scenes.
\newblock In \emph{EMNLP}.

\bibitem[{Krishna et~al.(2016)Krishna, Zhu, Groth, Johnson, Hata, Kravitz,
  Chen, Kalantidis, Li, Shamma, Bernstein, and Fei-Fei}]{krishnavisualgenome}
Ranjay Krishna, Yuke Zhu, Oliver Groth, Justin Johnson, Kenji Hata, Joshua
  Kravitz, Stephanie Chen, Yannis Kalantidis, Li-Jia Li, David~A Shamma,
  Michael Bernstein, and Li~Fei-Fei. 2016.
\newblock \href {https://arxiv.org/abs/1602.07332} {Visual genome: Connecting
  language and vision using crowdsourced dense image annotations}.

\bibitem[{Mao et~al.(2016)Mao, Huang, Toshev, Camburu, Yuille, and
  Murphy}]{mao2016generation}
Junhua Mao, Jonathan Huang, Alexander Toshev, Oana Camburu, Alan Yuille, and
  Kevin Murphy. 2016.
\newblock Generation and comprehension of unambiguous object descriptions.
\newblock In \emph{CVPR}.

\bibitem[{Matuszek()}]{matuszek2018grounded}
Cynthia Matuszek.
\newblock Grounded language learning: Where robotics and nlp meet.

\bibitem[{Mauceri et~al.(2019)Mauceri, Palmer, and Heckman}]{mauceri2019sun}
Cecilia Mauceri, Martha Palmer, and Christoffer Heckman. 2019.
\newblock Sun-spot: An rgb-d dataset with spatial referring expressions.
\newblock In \emph{Proceedings of the IEEE International Conference on Computer
  Vision Workshops}, pages 0--0.

\bibitem[{Murali et~al.(2019)Murali, Chen, Alwala, Gandhi, Pinto, Gupta, and
  Gupta}]{pyrobot2019}
Adithyavairavan Murali, Tao Chen, Kalyan~Vasudev Alwala, Dhiraj Gandhi, Lerrel
  Pinto, Saurabh Gupta, and Abhinav Gupta. 2019.
\newblock Pyrobot: An open-source robotics framework for research and
  benchmarking.
\newblock \emph{arXiv preprint arXiv:1906.08236}.

\bibitem[{Ralph and Moussa(2005)}]{ralph2005human}
Maria Ralph and Medhat Moussa. 2005.
\newblock Human-robot interaction for robotic grasping: A pilot study.
\newblock In \emph{2005 IEEE/RSJ International Conference on Intelligent Robots
  and Systems}, pages 454--459. IEEE.

\bibitem[{Shridhar and Hsu(2018)}]{shridhar2018interactive}
Mohit Shridhar and David Hsu. 2018.
\newblock Interactive visual grounding of referring expressions for human-robot
  interaction.
\newblock \emph{arXiv preprint arXiv:1806.03831}.

\bibitem[{Suchi et~al.(2019)Suchi, Patten, Fischinger, and
  Vincze}]{DBLP:conf/icra/SuchiPFV19}
Markus Suchi, Timothy Patten, David Fischinger, and Markus Vincze. 2019.
\newblock \href {https://doi.org/10.1109/ICRA.2019.8793917} {Easylabel: {A}
  semi-automatic pixel-wise object annotation tool for creating robotic {RGB-D}
  datasets}.
\newblock In  \cite{DBLP:conf/icra/SuchiPFV19}, pages 6678--6684.

\bibitem[{Yang et~al.(2019)Yang, Li, and Yu}]{yang2019dynamic}
Sibei Yang, Guanbin Li, and Yizhou Yu. 2019.
\newblock \href {https://doi.org/10.1109/ICCV.2019.00474} {Dynamic graph
  attention for referring expression comprehension}.

\bibitem[{Yang et~al.(2020)Yang, Li, and Yu}]{yang2020graph-structured}
Sibei Yang, Guanbin Li, and Yizhou Yu. 2020.
\newblock Graph-structured referring expressions reasoning in the wild.

\bibitem[{Yu et~al.(2016)Yu, Poirson, Yang, Berg, and
  Berg}]{referitgameeccv2016}
Licheng Yu, Patrick Poirson, Shan Yang, Alexander~C. Berg, and Tamara~L. Berg.
  2016.
\newblock Modeling context in referring expressions.
\newblock In \emph{Computer Vision -- ECCV 2016}, pages 69--85, Cham. Springer
  International Publishing.

\end{thebibliography}
\bibliographystyle{acl_natbib}
\clearpage
\appendix

In this Appendix, we provide the detailed setting of our grasping experiment environment in V-REP (Section A), as well as the real-world experiment using UR5 arm.

\section{Grasp Experiment in Simulator}
\label{sec:appendix-grasp-env}

\begin{figure}[t]
\begin{center}
\includegraphics[width=1.0\linewidth]{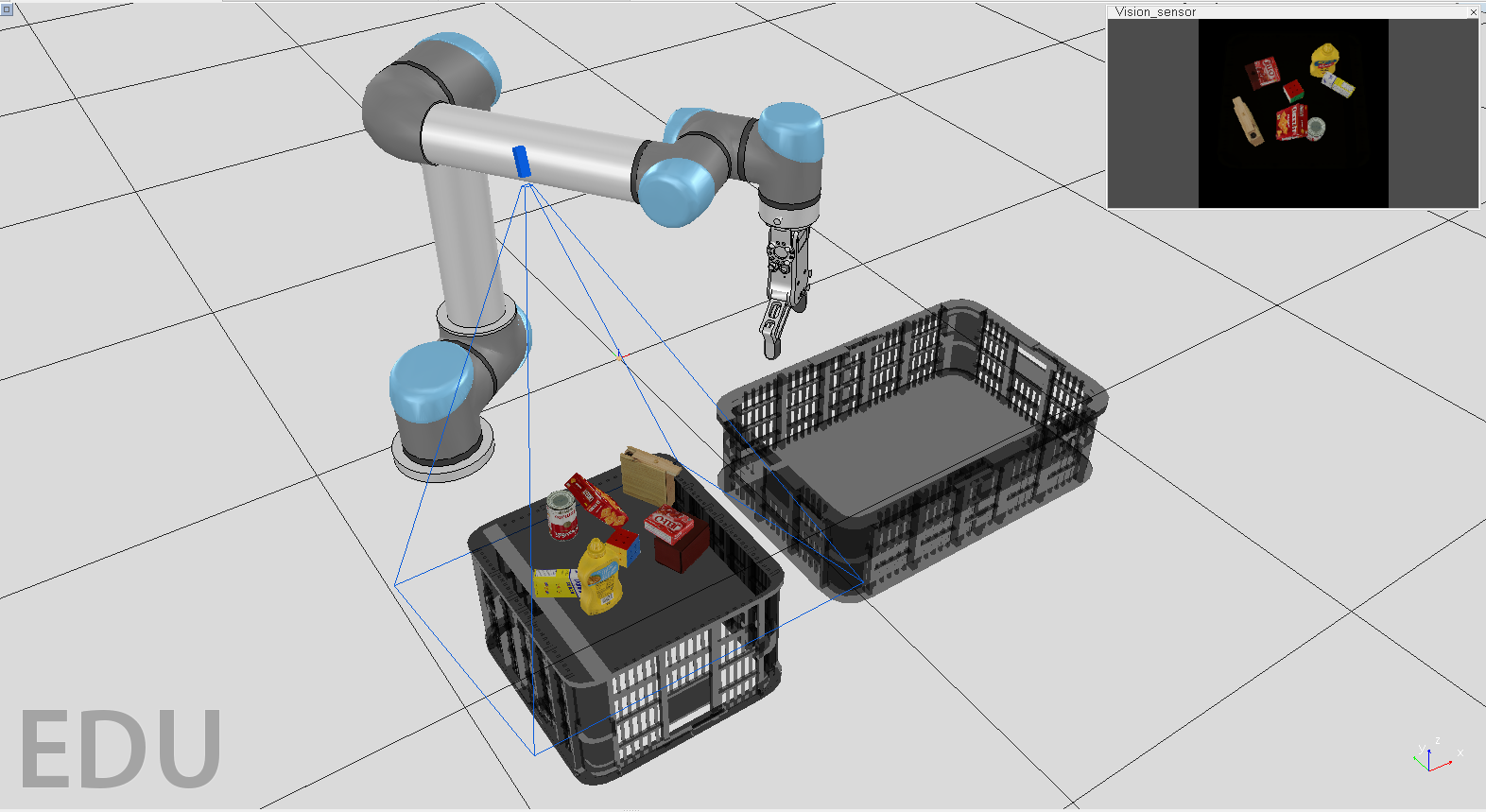}
\end{center}
\caption{\label{fig:grasp-env}
Grasp Experiment Environment.
}
\end{figure}

We use V-REP to set up the experimental environment of grasping tasks, which contains the UR5 arm, the RG2 gripper, a table, a box, the objects to grasp, and one depth camera. We show the grasp experiment environment in figure \ref{fig:grasp-env}. 

We produce the point cloud from the depth camera as the input and uses GDN \citep{jeng2020gdn} to find the grasps based on the point cloud. We perform a single object grasping experiment in three clutter levels, free, touching, stacked, with 4,6,8 objects in the scene, respectively. An object in the YCB \cite{calli2015ycb} set is randomly selected and placed on a table, and then the robot tries to grasp the object. If the robot successfully grasps the object from the table to the box, it counts one success. We do 11 trials and calculate the average success rate for each object.  

Table \ref{tab:grasp-exp} demonstrates the results of the single object grasping task in terms of success rate with 2D box and 3D Mask in different clutter level. Using a 3D mask rate than a 2D box as an input can get a higher average gripping success rate. The difference increases in a more dense cluttered scene, suggesting that with more accurate segmentation in 3D spatial environment is relatively unaffected in the cluttered and occluded environment.
\section{Real-World Experiment}
\label{sec:appendix-real-world-exp}
We use UR5 robotic arm to conduct the real-world experiment and equip Intel Realsense D415 to obtain the RGB and depth information. Additionally, we utilize  PyRobot \cite{pyrobot2019} to high-level interact with ROS Kinetic to control the robotic arm, and we adopt rapidly exploring random tree as our planning strategy to manipulate the movement of arm. For the real-world environment, we first set up a table with black table cloth, and put some objects on it. Then, ask the user to give a referring expression to the system, and it would identify the target object and grasp it.




\begin{table}
\centering
\begin{tabular}{ccccccc}
\hline
 & \multicolumn{3}{c}{\textbf{top}} & \multicolumn{3}{c}{\textbf{bottom}}\\
\textbf{}            & F   & T & S & F   & T & S \\
\hline
2D               & 65.9 & 40.9   & 28.4  & 72.7 & 51.5   & 21.6  \\
3D               & 72.7 & 48.5   & 36.4  & 75.0 & 56.1   & 31.8  \\
\hline
$\Delta$            & 6.8  & 7.6    & \textbf{8.0}   & 2.3  & 4.5    & \textbf{10.2} \\
\hline
\end{tabular}
\caption{\label{tab:grasp-exp}
2D (box) and 3D (mask) Grasp Experiment on different clutter level (\%), F: Free, T: Touching, S: Stacked.
}
\end{table}

\end{document}